\documentclass[lettersize,journal]{IEEEtran}
\usepackage{amsmath,amsfonts}
\usepackage{algorithmic}
\usepackage{algorithm}
\usepackage{array}
\usepackage[caption=false,font=normalsize,labelfont=sf,textfont=sf]{subfig}
\usepackage{textcomp}
\usepackage{stfloats}
\usepackage{url}
\usepackage{verbatim}
\usepackage{graphicx}
\usepackage{cite}
\usepackage{amssymb}  % assumes amsmath package installed
\usepackage{booktabs}

\usepackage[utf8]{inputenc}

\usepackage{glossaries}
\makeglossaries

\newacronym{epdms}{EPDMS}{Extended Predictive Driver Model Score}
\newacronym{matryoshka}{Matryoshka}{}
\newacronym{navsim}{NAVSIM}{}
\newacronym{pdm}{PDM}{Predictive Driver Model}
\newacronym{sae}{SAE}{Sparse Autoencoder}

\newglossaryentry{topk}{
    name={TopK},
    description={Top-$k$ selection}
}

\begin{document}

% \title{Driving the Wrong Way: Correcting End2End Autonomous Driving Models with Interpretability}
\title{Driving the Wrong Way: Leveraging Interpretability in End2End Autonomous Driving Models}

\author{Franz Motzkus,
\thanks{Franz Motzkus is with AUMOVIO, Germany, and with the Department of Applied Computer Science, University of Bamberg, Germany.}% <-this % stops a space
% 96047 Bamberg, 
Sebastian Bernhard
% \thanks{Sebastian Bernhard is with the Technical University of Applied Sciences Augsburg, Germany.}
}

% \markboth{Transactions on Intelligent Transportation Systems,~Vol.~XX, No.~X, Month~2026}%


\maketitle

%%%%%%%%%%%%%%%%%%%%%%%%%%%%%%%%%%%%%%%%%%%%%%%%%%%%%%%%%%%%%%%%%%%%%%%%%%%%%%%%
\begin{abstract}
The increasing adoption of end-to-end learning for autonomous driving introduces increased model complexity and opacity, raising the risk of learning undesired or erroneous behavior.
In this work, we integrate unsupervised dictionary learning as a post hoc interpretability module within state-of-the-art driving models to decompose driving behavior into semantically meaningful concepts while demonstrating their causal influence on the model's driving decisions.
We propose a stepwise framework for extracting and interpreting meaningful concepts from the end-to-end model and connecting them to the multifaceted model outputs, thereby revealing the underlying decision-making logic for the prediction of future trajectories.
Furthermore, targeted interventions at the concept level allow us to manipulate and correct driving decisions, resulting in measurable improvements in overall driving performance.
We thus demonstrate how interpretability can effectively be used to reduce model opacity, uncover erroneous behavior, and enable targeted mitigation, ultimately boosting model performance.

\end{abstract}


%%%%%%%%%%%%%%%%%%%%%%%%%%%%%%%%%%%%%%%%%%%%%%%%%%%%%%%%%%%%%%%%%%%%%%%%%%%%%%%%

\section{INTRODUCTION}

End-to-end learning has recently emerged as a prominent approach for autonomous driving systems.
In contrast to conventional modular pipelines with explicitly separated perception, prediction, and planning components, modern approaches integrate multimodal inputs, including camera recordings and top-level commands, into unified transformer-based architectures~\cite{chen_end_2024}.
These systems jointly learn perception, temporal reasoning, and decision-making, achieving state-of-the-art performance on open-loop benchmarks such as \gls{navsim}~\cite{dauner_navsim_2024,wei_pseudo_2025}.

However, this architectural unification comes at the cost of reduced transparency, as the model complexity increases and clear interfaces that previously enabled module-level testing disappear.
Consequently, internal decision-making becomes increasingly opaque, making the analysis of failure modes more difficult and resulting in expensive error debugging through repeated data curation and retraining cycles guided mainly by best guesses.
Especially in terms of safety and interpretability, the field of end-to-end autonomous driving lacks generalizing solutions for holistically describing model behavior and offering comprehensive insights into failure modes~\cite{kuznietsov_explainable_2024,atakishiyev_safety_2025}.

\begin{figure*}
    \centering
    \includegraphics[width=\linewidth]{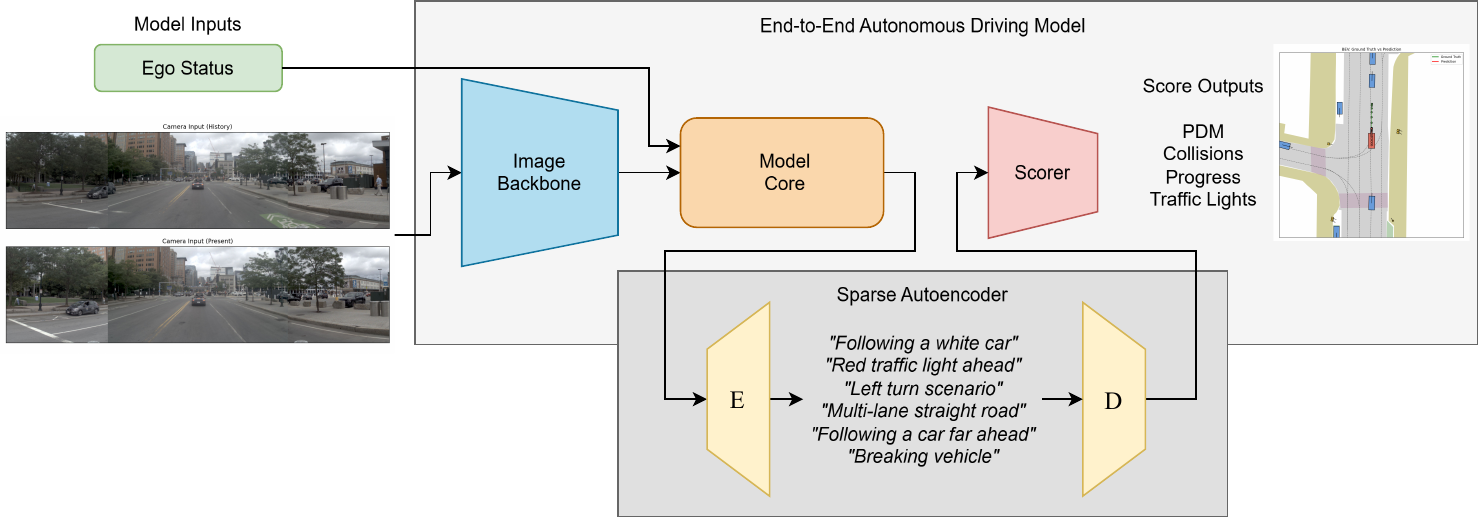}
    \caption{Framework Architecture for Integrating an Interpretability Layer into End-to-End Autonomous Driving Models}
    \label{fig:framework}
\end{figure*}

In this work, we address this limitation by introducing a concept-based interpretability framework for end-to-end autonomous driving.
Our approach reveals interpretable latent-space components and directly connects them to trajectory-level decision scores.
Specifically, we integrate dictionary learning via \glspl{sae} into the model's feature space, decomposing latent activations into sparse, human-interpretable concepts.
Rather than treating the latent space as an opaque collection of convoluted features, this decomposition provides a more transparent representation in which individual concepts correspond to meaningful patterns learned by the model.
Consequently, complex driving decisions can be expressed as a composition of semantically meaningful elements with causal influence on the predicted future trajectory.
This concept-based perspective offers a more interpretable view of the model's internal reasoning process while preserving the representational capacity required for autonomous driving.

To validate our approach, we perform concept-level interventions by selectively suppressing individual latent components.
We show that interventions on concepts associated with erroneous behavior lead to measurable improvements in downstream driving metrics, including collision-avoidance, drivable area and traffic light compliance.
This indicates that the identified concepts are not merely correlational artifacts but functionally relevant components of the model's decision-making.
The results further demonstrate how increased interpretability can be leveraged for targeted model refinement and improved overall performance.

Our contributions are the following:
\begin{enumerate}
    \item We present the first structured and generalized approach for interpreting driving decisions in end-to-end autonomous driving models, providing concept-level explanations for the scoring of individual candidate trajectories.
    \item We introduce a concept-based decomposition of latent representations that identifies the most influential factors within individual driving scenarios, enabling fine-grained attribution of model decisions.
    \item We establish a direct correspondence between learned concepts and the predicted quality scores for trajectories, revealing which concepts are influencing the distinct driving behaviors such as collision avoidance, traffic-light compliance and progress.
    % or even to individual subscores with diverging semantic meaning (lane, traffic lights, progress)
    % This provides detailed insights into which latent factors contribute to specific driving capabilities and failure modes.
    \item We demonstrate that the identified concepts can be leveraged for targeted model adaptation and manipulation, enabling improvements in downstream driving performance without the need for retraining.
\end{enumerate}

\section{RELATED WORK}

\subsection{End-to-End Autonomous Driving}

With the adoption of the end-to-end learning paradigm in autonomous driving AI models, multiple different approaches from imitation learning to reinforcement learning have been introduced~\cite{chen_end_2024}.
\emph{Imitation learning} methods train from (human) demonstrations via behavior cloning, mimicking the ground truth expert's driving decisions from real-life driving recordings.
Popular state-of-the-art models using imitation learning are 
Hydra-MDP~\cite{li_hydra_2024} scoring on a trajectory vocabulary, and Emma~\cite{hwang_emma_2025} incorporating a language model to build a multimodal predictor. 
% TransFuser~\cite{chitta_transfuser_2023}, Think Twice~\cite{jia_think_2023}, UniAD~\cite{hu_planning_2023}, and Hydra-MDP~\cite{li_hydra_2024}.
Typically, they output trajectories with estimated quality scores and the training objective is to approximate the ground truth trajectory from a real-world recording.

\emph{Reinforcement learning} instead typically operates in a virtual environment such as a simulator, where the model's interaction with the environment is optimized based on a reward function.
However, the gap between autonomous driving simulation and real-world driving is still a foundational problem.

A group of current approaches is referred to as offline reinforcement learning, where training is based on recordings with ground-truth driving behavior, but alternative trajectories generated by the model can also be evaluated and optimized~\cite{dauner_navsim_2024}. 
By generating possible next steps of the scenario, the influence of the environment interaction is even enhanced~\cite{wei_pseudo_2025}.
These models typically predict trajectories with an estimated \gls{pdm} score that incorporates an imitation score and additional penalty scores, including collision, drivable area, and traffic light compliance, along with progress along the ground-truth trajectory.

\cite{chen_vadv2_2024} introduces a probabilistic scoring over a predefined dictionary of trajectories.
Later approaches adapt the \gls{pdm} scoring for a quality assessment of trajectories.

\textbf{Hydra-MDP}~\cite{li_hydra_2024} trains multiple diverse trajectory heads and selects among them at inference time, explicitly addressing the multimodality of human driving behavior to improve robustness under the \gls{pdm} evaluation protocol.

\textbf{GTRS}~\cite{li_generalized_2025} builds upon the Hydra-MDP model and introduces an additional trajectory generation step via a diffusion model. Instead of a fixed-trajectory dictionary, possible trajectories are generated iteratively.

\textbf{iPad}~\cite{guo_ipad_2025} shifts from dense BEV grid features to an iterative proposal-centric paradigm. A set of sparse, learnable trajectory proposals serves as the central organising principle: its ProFormer encoder iteratively refines proposals and their associated BEV queries through proposal-anchored attention, concentrating feature extraction on planning-relevant regions. 
Lightweight proposal-centric auxiliary tasks for mapping and collision prediction further improve planning quality.

\subsection{Interpretability in End-to-End Autonomous Driving}

While the end-to-end training schema drastically increases the opacity of the presented models compared to classical approaches with AI-based perception and rule-based planning, multiple approaches seek to incorporate interpretability to (parts of) the model~\cite{atakishiyev_safety_2025}.

\emph{Saliency-based methods}~\cite{bojarski_explaining_2017,bojarski_visualbackprop_2018} inspect model predictions through saliency, and gradient-based attribution (e.g.\ GradCAM, Integrated Gradients). 
These methods are widely used due to their simplicity but are limited to identifying \textit{where} the model focuses within the input, providing no account of \textit{what} the model has learned to represent internally.
Especially in autonomous driving, where multiple factors influence the driving decision, saliency maps are limited as they highlight only the most prominent aspect.

\emph{Attention visualization methods} use the models' built-in attention mechanisms to highlight the most-attended information in the front camera view~\cite{kim_interpretable_2017,sauer_conditional_2018,liu_interpretable_2020,araluce_leveraging_2024} or in an abstracted birds-eye-view~\cite{Kim_2020_CVPR_Workshops,holtz_what_2026}.
Another work optimizes the model itself to yield diverse feature maps in input space~\cite{Mirzaie_interpretable_2025}.

\emph{Latent space interpretability:}
While attention visualization reveals highly attended information, other latent space interpretability methods are rarely applied to end-to-end autonomous driving models. 
\cite{morton_simultaneous_2017} uses latent space representations of an LSTM network as additional input for predicting driving actions in a policy network.
\cite{tian_deeptest_2018} introduces a testing framework using neuron coverage, thereby testing the influence of input perturbations on intermediate representations.
Neither method focuses on making the latent space more interpretable.

In~\cite {hu_stp3_2022}, an end-to-end autonomous driving model is specifically designed to have interpretable intermediate representations between perception, prediction and planning, yielding separate scene representations for drivable area, lane markings and traffic participants. 
However, the latent space representations are rather controlled with the necessity of using the specifically introduced architecture.

\emph{Vision–language models and vision–language–action models (VLM/VLA)} present an emerging research direction in autonomous driving~\cite{zhou_autovla_2025, zhou_opendrivevla_2026}.
While these approaches open promising directions for leveraging interpretability in autonomous driving models, more in-depth analyses at the level of underlying concepts---through which the different aspects of driving decisions can be disentangled and examined---are still largely missing.
To our knowledge, no other concept-based latent space interpretability approach for end-to-end autonomous driving models exists.

\section{PRELIMINARIES}

\subsection{Sparse Autoencoder Learning}

Sparse Autoencoders have first shown to extract interpretable feature directions in language models~\cite{templeton2024scaling}, but also proven useful in vision models~\cite{daujotas_interpreting_2024, lim_sparse_2025} or even trajectory prediction~\cite{motzkus_trust_2025}.
The latent space of AI models is typically highly entangled, with single neurons encoding multiple unrelated features, making them non-interpretable.
This polysemanticity of neurons hinders a mechanistic understanding of individual features~\cite{elhage_superposition_2022}, in which the functioning of single network units is explored.
As neural networks often learn more features than having dimensions, this superposition can be resolved with sparse autoencoders~\cite{bricken_monosemanticity_2023}.
By expanding the latent space while enforcing sparse representations, they provide a principled approach to decomposing an entangled representation into linear directions corresponding to interpretable, monosemantic features, denoted as concepts.

Formally, given model activations $x \in \mathbb{R}^d$, an overcomplete dictionary $D \in \mathbb{R}^{d\times m}$ with $m > d$ is learned, such that $x \approx Dz$ with $z \in \mathbb{R}^m$ forming a sparse representation.
\glspl{sae} jointly learn the dictionary $D$ and a sparse encoder $E: \mathbb{R}^d \to \mathbb{R}^m$  by minimizing the reconstruction loss $\|x - D E(x)\|_2^2$ between the activation vector $x$ and it's reconstruction  $D E(x)$.
To achieve sparse representations in the latent codes $z$, sparsity  objectives like a $\ell_1$-regularization or a top-$k$ activation retaining only the $k$ highest values in $z$ are applied~\cite{huben_sparse_2024, gao_scaling_2025}.
Multiple works adapt the standard \gls{sae} training scheme to achieve a different feature distribution.

\emph{\gls{matryoshka}} training~\cite{bussmann_learning_2025} builds on the idea of nested representations, where subsets of latent dimensions define a sequence of progressively more expressive autoencoders. 
Given a latent code $z \in \mathbb{R}^m$, whose coordinates correspond to $m$ latent features, we consider prefixes 
$z_{1:m_1}, z_{1:m_2}, \dots, z_{1:m}$ with $m_1 < m_2 < \dots < m$.
Each subset of latent features defines its own reconstruction, yielding a sequence of valid autoencoders with increasing reconstruction fidelity and inducing a hierarchy in which smaller subsets capture general features, while larger subsets include progressively more specific ones.

\emph{Archetypal} \glspl{sae}~\cite{fel_archetypal_2025} introduce additional convexity constraints on the learned dictionary.
Let $X \in \mathbb{R}^{d \times n}$ 
denote a set of $n\in\mathbb{N}$ reference data points, where each column corresponds to an input 
in $\mathbb{R}^d$. The dictionary is parameterized as
$ D = X A$,
where $A \in \mathbb{R}^{n \times m}$ satisfies $A \ge 0$ and 
$\mathbf{1}^\top A = \mathbf{1}^\top$, such that each dictionary atom 
$d_j = X a_j$ is a convex combination of data points.
Thus, in contrast to standard \glspl{sae} with unconstrained dictionaries, latent features correspond to archetypes that lie in the convex hull of the data.
In practice, these archetypes can be compared to representative data prototypes and resemble cluster-like summaries of the dataset, with each dictionary atom capturing a convex combination of similar inputs.

\subsection{SAE Evaluation}
We evaluate \glspl{sae} in two categories, whereas the first one is the reconstruction fidelity of the learned representations, while the other rates the alignment with desirable features based on our use case.

The reconstruction quality evaluates how well the SAE preserves the information contained in the original latent embeddings.
The cosine similarity measures the alignment between the models latent space representation $x$ and its reconstruction after the SAE $\hat{x}$ as
$
    \mathrm{cos}(x, \hat{x}) = 
    \frac{x^\top \hat{x}}{\|x\|_2 \, \|\hat{x}\|_2},
$
It captures the preservation of directional information in latent space.

The explained variance determines the fraction of variance in $x$ explained by $\hat{x}$ and is given by
$
    \mathrm{EV} = 1 - \frac{\mathrm{Var}(x - \hat{x})}{\mathrm{Var}(x)},
$
where $\mathrm{Var}(\cdot)$ denotes the empirical variance over the dataset.

As the investigated end-to-end models include an ego status, consisting of values for velocity, acceleration and a top-level driving direction, into their computation, we assume it to be represented in the \gls{sae} with individual features.
As a quality criteria, we can thus compute how well the ego status is represented in the \gls{sae} neurons.
For each ego-status variable $e_j$ (velocity, acceleration, driving command), we compute the Pearson correlation $r_{n,j}$ between every alive \gls{sae} neuron $i$ and $e_j$ across the dataset. 
The ego correlation score is defined as:
\begin{equation}
    S_{\text{corr}} = \frac{1}{|\mathcal{E}|} \sum_{j \in \mathcal{E}} \max_i |r_{i,j}|
\end{equation}
where $\mathcal{E}$ is the set of non-constant ego features. 
A score of 1 indicates that every ego variable is perfectly linearly represented by at least one SAE neuron.

In the ego probing, we train linear probes on the \gls{sae} codes to predict ego-status variables, using a log-based train/test split (entire driving logs are held out) to prevent temporal leakage. 
Three probes are evaluated: (1) a logistic regression classifier for the 
driving command (left/straight/right), yielding test macro-F1 $F_{\text{cmd}}$; 
(2) a Ridge regressor for continuous ego dynamics (velocity, acceleration), 
yielding joint test $R^2_{\text{ego}}$; and (3) a scenario-ID classifier measuring log memorization, with accuracy $a_{\text{sc}}$. The composite probing score is:
\begin{equation}
    S_{\text{probe}} = 0.5 \, F_{\text{cmd}} + 0.5 \, R^2_{\text{ego}} 
    - 0.2 \max(0,\; a_{\text{sc}} - a_{\text{random}})
\end{equation}
The scenario penalty discourages representations that merely memorize log identity rather than encoding generalizable driving concepts.

\subsection{Circuit Discovery}
\label{preliminaries:circuits}

Mechanistic interpretability~\cite{saphra_mechanistic_2024} refers to the analysis of single parts of a frozen neural network to extract human-understandable components.
In mechanistic interpretability, circuit analysis uncovers minimal computational subgraphs between neurons modeling specific model behavior.
With the introduction of \glspl{sae}, circuits no longer operate on polysemantic neurons, but directly on the monosemantic concept level~\cite{marks_sparse_2025}, amplifying their strength in revealing interpretable, human-auditable subgraphs.

Multiple different algorithms for the computation of circuits exist varying in the trade-off between computation cost and accuracy.
As a computation based on the removal of single neurons and a following full network evaluation is computationally not feasible, we apply a circuit computation based on neuron attributions.
Given a set of \gls{sae} latent codes $Z \in \mathbb{R}^{n \times m}$ and per-sample attribution scores $\alpha_{h,i}$, we identify a sparse circuit of \gls{sae} neurons causally mediating each planning head output through a four-stage pipeline.

\paragraph{Attribution Patching.}
For each sample latent $z$ in $Z$ and head $h$ of the model's $H$ outputs, the influence of neuron $i$ is estimated via the gradient-weighted activation difference:
\begin{equation}
    I^{\text{attr}}_{h,i} = \overline{|\alpha_{h,i}|} \cdot \mathrm{sign}(\bar{\alpha}_{h,i}),
    \qquad
    \alpha_{h,i} = \frac{\partial f_h(z)}{\partial z_i} \cdot (z_i - \bar{z}_i)
\end{equation}
where $\bar{z}_i$ and $\bar{\alpha}_{h,i}$ denote averages over the dataset $Z$.
This yields a signed influence matrix $\mathbf{I}^{\text{attr}} \in \mathbb{R}^{H \times m}$.
This provides a fast, approximate estimate of the causal influence of all latent features on each head simultaneously.

\paragraph{Activation Patching.}
For the top-$2q$ candidate features selected by attribution patching 
(where $q$ denotes the target circuit size), we perform 
exact causal interventions. 
Each neuron $i$ is individually patched to its dataset-mean baseline $\bar{z}_i$, and the causal effect on head $h$ is measured as:
\begin{equation}
    I^{\mathrm{act}}_{h,i} = \mathbb{E}_z \left[
        \left| f_h(z) - f_h\!\left(z^{(i \leftarrow \bar{z}_i)}\right) \right|
    \right]
\end{equation}
where $f_h(z)$ denotes the output of head $h$ as a function of the latent 
representation $z$.

\paragraph{ACDC Pruning.}
The influence matrix is sparsified by retaining only edges exceeding a relative 
threshold $\tau$ per head:
\begin{equation}
    \mathcal{E} = \left\{
        (h, i) \;\Big|\;
        \frac{I^{\mathrm{act}}_{h,i}}{\max_{i'} I^{\mathrm{act}}_{h,i'}} > \tau
    \right\}
\end{equation}
This yields the final sparse circuit 
$\mathcal{E} \subseteq \{1,\ldots,H\} \times \{1,\ldots,m\}$.

\paragraph{Neuron Specificity.}
Each neuron's degree of head-specificity is quantified via the Gini coefficient 
over its normalized influence distribution across heads:
\begin{equation}
    g_i = 1 - \frac{2}{H-1} \sum_{k=1}^{H-1} C_k(i),
    \qquad
    C_k(i) = \frac{\sum_{j \leq k} \tilde{I}_{j,i}}{\sum_j \tilde{I}_{j,i}}
\end{equation}
where $\tilde{I}_{j,i}$ are the normalized influences sorted in ascending order. $g_i = 1$ indicates a neuron exclusively influencing a single head; $g_i = 0$ indicates uniform influence across all heads.

\section{METHODOLOGY}
\label{sec:methodology}

To uncover the factors driving high-stakes decisions in end-to-end autonomous driving models, we propose a structured, multi-step interpretability framework.
We begin by using \glspl{sae} to disentangle the latent space of an end-to-end model into separate, interpretable feature directions.
This provides a lens into otherwise opaque internal representations without significantly altering the original predictive performance.
% Next, these learned feature directions are systematically linked to semantics, providing the connection between model encoding and human understanding.
Next, we systematically assign semantic meaning to these learned feature directions, establishing a connection between abstract model encodings and human-understandable concepts.
Building on this, we analyze individual driving scenarios by decomposing model decisions into their underlying feature contributions.
Additionally, feature activations are traced back into the input space, producing camera-based attribution maps that localize the corresponding features in each scene and provide another level of interpretability.
In the subsequent step, we use circuit analysis to connect these semantic features to the model's outputs.
This enables us to directly quantify how individual features contribute to different driving properties, e.g. by inspecting which features are relevant for predicting the traffic light compliance.
Finally, we leverage these insights to identify failure cases and undesired behaviors, enabling targeted mitigation strategies and improving the overall reliability of the system.

\subsection{Selecting an Expressive Latent Space}

As the first step of our framework, a reasonable latent space representation for injecting the \gls{sae} needs to be identified within the model.
The criteria are as follows:
\begin{itemize}
    \item The latent space may encode a dense summary of the scene understanding, including all model inputs as observed by the model.
    \item The model should base the prediction on the encoded scene understanding of this latent space, so that the trajectory scoring can be linked to the extracted \gls{sae} features.
    \item The dimensionality of the latent space should ideally be of moderate size (i.e., $ d=256$) and allow for a 1D vector representation (per trajectory).
\end{itemize}

In the chosen models, we choose the latent space representation directly preceding the scoring module, yielding a 1D representation per trajectory proposal.
This representation is particularly suitable, as it contains the complete scene understanding while its features are directly inform the model scoring. 

% \subsection{Training the Sparse Autoencoder}
\subsection{Learning Interpretable Feature Directions}

In the second step, we train an \gls{sae} on the selected latent space to disentangle it into sparse, interpretable features.
The \gls{sae} is trained on the frozen end-to-end model's latent representations, minimizing the reconstruction error:
\begin{equation}
\mathcal{L}_{\text{rec}} =
\frac{1}{N}\sum_{p=1}^{N}
\left\| x_p - \hat{x}_p \right\|_2^2.
\end{equation}

Sparsity is enforced by a top-k activation function on the \gls{sae} latent space activation. To prevent inactive concepts, we introduce an additional reanimation term:
\begin{equation}
\mathcal{L}_{\text{reanim}} =
- \lambda_r
\sum_{i \in \mathcal{D}}
\frac{1}{N}\sum_{p=1}^{N} h^{\text{pre}}_{p,i},
\end{equation}
where $\mathcal{D}$ denotes neurons inactive for a predefined number of steps and $\lambda_r>0$ denotes the scaling factor.
The term $h^{\text{pre}}_{p,i}$ describes the \emph{pre-activation} of latent feature $i$ for sample $p$, i.e., the value before the application of the top-$k$ sparsifying nonlinearity.

% \subsection{Neuron Interpretability}
\subsection{Assigning Semantic Meaning to Features}
\label{sec:semantics}
Having obtained disentangled, monosemantic feature directions with the \gls{sae}, the third step focuses on assigning semantic meaning to individual SAE neurons.
In this context, semantic refers to a consistent, human-understandable concept that can be expressed in natural language.
We use the following strategies for revealing neuron semantics.

\subsubsection{Activation/Attribution Maximization}
Activation and attribution maximization~\cite{olah_feature_2017,achtibat_attribution_2023} are data-based approaches that extract samples from a reference dataset that maximally activate the specified neuron or produce the highest attribution, meaning the neuron is maximally used for the model's prediction in that sample.
Formally, given a neuron $i$ and a reference dataset $X$, the top-$k$ maximally activating samples are retrieved as:
\begin{equation}
    \mathcal{S}^*_i = \underset{\mathcal{S} \subseteq X,\, |\mathcal{S}|=k}{\arg\max}
    \sum_{x \in \mathcal{S}} z_i(x)
\end{equation}
where $z_i(s)$ is the activation of neuron $i$ on sample $x$. For attribution maximization, $z_i(x)$ is replaced by the neuron's attribution score with respect to the model output. 
As explained in the next section, it captures not merely whether a neuron fires but whether it contributes to the prediction.

\subsubsection{SAE Neuron Attribution}
\label{sec:saeattr}
Input attribution methods highlight input units based on their importance towards a model output objective.
We employ Concept Relevance Propagation (CRP)~\cite{achtibat_attribution_2023}, which builds on Layer-wise Relevance Propagation (LRP)~\cite{bach_pixelwise_2015}.
LRP is a modified gradient backpropagation method, using specific rules to preserve the semantic signal from an output objective in the gradient backpropagation. 
CRP extends the gradient pass with an intermediate gradient masking to yield the gradient/attribution signal for a single neuron.
The use of LRP is particularly advantageous in this context, as its relevance conservation property enables precise attribution of the prediction to individual latent features, while its robustness to gradient noise facilitates effective filtering of meaningful semantic signals compared to standard gradient-based methods.

In our setting, the model assigns a scalar score $f(x_t)$ to the latent activation $x_t \in \mathbb{R}^d$ of each trajectory proposal. 
The score of the selected trajectory is used to initialize the relevance propagation. 
Starting from this scalar output, relevance is propagated backward through the network to the \gls{sae} latent space with its representation $z_t = E(x_t) \in \mathbb{R}^m$.

During propagation, relevance is distributed at each layer proportionally to the contribution of each neuron:
\begin{equation}
R_i =
\frac{z_i w_i}{\sum_j z_j w_j}
R_{\text{out}},
\end{equation}
where $w_i$ denotes the contribution weight to the layer output, and $R_{\text{out}}$ is the relevance from the subsequent layer.
This yields relevance scores $R_i$ at the level of \gls{sae} neurons, quantifying their contribution to the selected trajectory score.
The \gls{sae} neurons are then ranked according to their relevance scores $R_i$, allowing us to identify the concepts most responsible for the selected trajectory.

To isolate the contribution of a single neuron $i$, we apply concept masking at the \gls{sae} layer during relevance propagation:
$R^{(i)} = R \odot e_i$,
% $z_p^{(i)} = z_p \odot \mathbf{e}_i$,
%
where $e_i$ is the unit vector.
The masked relevance is then propagated back to the model inputs and reveals highly attributed visual patterns in the camera input or the ego status corresponding to that neuron.

% \subsection{Circuit Analysis}
\subsection{Linking Features to Model Outputs}
\label{sec:circuits}
In the fourth step, we connect the \gls{sae} features to the model predictions using circuit analysis as described in Section~\ref{preliminaries:circuits}.
For end-to-end autonomous driving models predicting the \gls{pdm} subscores for a set of trajectory proposals, the individual components of the \gls{pdm} score can be linked to the extracted features.
Depending on the user's interest, a subset of \gls{sae} neurons and/or a selection of \gls{pdm} elements can be selected for the circuit computation.
The resulting circuit represents a sparse mapping between latent features and model outputs, identifying which features causally influence specific components of the prediction. 
The user can thus comprehend how the interpretable features influence the component scores.

% \subsection{Neuron Manipulation}
\subsection{Manipulating the Model Logic}
\label{sec:manipulation}
Finally, we leverage the established feature-to-output relationships to identify failure cases and enable targeted feature-level interventions.
Recalling that an \gls{sae} neuron encodes a distinct semantic feature direction, its activation can be easily controlled by globally amplifying or attenuating this feature across all samples.
Suppose the model mistakenly learned a spurious feature that undesirably influences the trajectory scoring.
Since the feature describes a distinct direction in the latent space, it is likely captured and encoded in an \gls{sae} neuron.
The influence of this spurious feature can then be eliminated simply by zeroing the activation of that neuron, avoiding the need for costly dataset adjustments or retraining.

For models that predict the individual components of the \gls{pdm} score with separate heads, the relationship between \gls{sae} neurons and specific score components becomes directly accessible.
Strategically placing the \gls{sae} module allows targeted interventions in the causal relations modifying which neurons influence which scoring heads.
Note that the patterns learned from the data likely induce strong cross-associations between features and prediction heads that are driven by correlation rather than true causality.

\section{RESULTS}

\begin{table*}[ht]
\centering
\begin{tabular}{lllcccccc}
\toprule
Version & \#Neurons & $k$ & Dead Neurons (\%) $\downarrow$ & Cosine Similarity $\uparrow$ & Explained Variance $\uparrow$ & Ego Corr $\uparrow$ & Ego Probing $\uparrow$ & EPDMS $\uparrow$\\
\midrule
\gls{topk} & 64 & 16 & 7.8 & 0.9824 & 0.9651 & 0.7097 & 0.8226 & 0.485 \\
\gls{topk} & 128 & 32 & 8.6 & 0.9904 & 0.9808 & 0.7568 & 0.8451 & 0.418 \\
\textbf{\gls{topk}*} & 256 & 64 & \textbf{0.0} & \underline{0.9975} & \underline{0.9951} & \textbf{0.8454} & \underline{0.9104} & 0.497\\
\gls{topk} & 1024 & 64 & 6.9 & 0.9950 & 0.99 & \underline{0.8010} & 0.8932 & 0.51 \\
\gls{topk} & 1024 & 128 & 4.5 & 0.9962 & 0.9923 & 0.7596 & 0.9031 & \textbf{0.552} \\
\midrule
Matryoshka & 64 & 16 & \textbf{0.0} & 0.0907 & 0.0126 & 0.4749 & 0.7603 & 0.08\\
Matryoshka & 128 & 32 & 19.5 & 0.9904 & 0.9809 & 0.5602 & 0.8409 & 0.441 \\
Matryoshka & 256 & 64 & 23.8 & 0.9949 & 0.9898 & 0.6151 & 0.8821 & 0.494\\
Matryoshka & 1024 & 64 & 9.0 & 0.9952 & 0.9903 & 0.7214 & 0.8977 & 0.508 \\
Matryoshka & 1024 & 128 & 16. & \textbf{0.9979} & \textbf{0.9957} & 0.7851 & \textbf{0.9105} & 0.524 \\
\midrule
Archetypal & 1024 & 128 & 16.3 & 0.9392 & 0.8726 & 0.5318 & 0.4439 & 0.151 \\
MatryArch & 1024 & 128 & 22.1 & 0.9538 & 0.9054 & 0.5364 & 0.6402 & 0.238 \\
\bottomrule
\end{tabular}
\caption{Performance Evaluation for multiple trained Sparse Autoencoders with differing settings on the GTRS model.}
\label{tab:gtrs_trained_saes}
\end{table*}

\subsection{Selecting the right SAE Parametrization}

For the first two steps of our framework (Section~\ref{sec:methodology}), we train multiple \glspl{sae} with varying architectures and hyperparameters on latent representations of the GTRS and iPAD model and measure their performance across several metrics.
The number of active SAE neurons ($\#neurons - \#dead\_neurons$, where dead neurons denote units that never activate across the dataset) determines the number of learned features.
The reconstruction quality of the model-internal embeddings---i.e., how well the SAE can reproduce the original latent representations---is determined by the cosine similarity and explained variance. 

To evaluate interpretability with respect to problem-specific information, we introduce two metrics.
Ego correlation measures the alignment of ego status features (velocity, acceleration, high-level direction) to the \gls{sae} neurons, while ego probing is a combined score testing the identifiability of ego status features in the \gls{sae} neurons while simultaneously penalizing the discriminability of individual scenarios (overfitting to scenarios).
Finally, the \gls{epdms} estimates the downstream performance with the \gls{sae} injected into the model.
The performance results are summarized in Table~\ref{tab:gtrs_trained_saes}.

While \glspl{sae} in language models typically require a high expansion factor, we experience substantially less required neurons for vision-based models.
In this case, it needs to be considered, that the inputs are derived from a limited number of scenarios, so that a pure scenario learning might occur, when choosing the latent space expansion too high.
The critical task is to find the right parametrization enabling a level of abstraction with meaningful features being encoded in the \gls{sae} neurons and with the performance not being degraded too much.
At first, the best-fitting \gls{sae} architecture needs to be determined.

In our experiments, archetypal-based \glspl{sae} substantially perform worse than the \gls{topk} and the \gls{matryoshka} version.
Inspecting the ego correlation and ego probing, we assume that the archetypal training rather tends to strengthen memorization effects as the \gls{navsim} data naturally includes multiple scenes per scenario, recorded at different time steps, or even includes scenes in the same location at different dates.
The data is thus strongly correlated, which easily reflects in the pre-computed clusterings that archetypal \glspl{sae} build upon.
In particular, archetypal SAEs ground their representations in dominant patterns of the data distribution, similar to cluster centroids or archetypes, such that individual neurons correspond to prototypical feature combinations. 
This reliance on representative data modes can amplify scenario-specific correlations, as frequently co-occurring features within a scenario are captured jointly and reinforced in the learned archetypes.

The \gls{topk} and \gls{matryoshka} \glspl{sae} perform significantly better, whereas several differences between both variants can be seen.
The ratio of dead neurons is comparably low for all\gls{topk} variants, while \gls{matryoshka} rather under-utilizes the available dictionary size by producing more dead neurons, and thus provides less learned features.

Interestingly, most of the other scores are quite similar between both variants and vary rather based on the latent size.
Only the ego correlation hints at a better separation of features related to the ego status in the \gls{topk} variants, presenting them as the best-fitting version.

Considering the optimal size for the \gls{sae} latent space, we observe the following.
Larger latent spaces enable the model to better recreate the model embedding, as more details can be encoded with a higher number of neurons.
Especially the downstream model performance is better with larger \glspl{sae}.

However, when inspecting the other metrics, the reconstruction does not necessarily improve from a certain neuron number.
Especially the medium sized \gls{topk} variant yields the best reconstruction.
The difference in performance is most apparent in the ego correlation and ego probing scores.
With a higher number of neurons, the relevant ego status features rather get distributed across neurons, hindering a consistent, disentangled encoding.
With a higher number of neurons, the probability of (re-)learning memorization effects also increases.

Compressing the \gls{sae} latent space too much lowers the reconstruction and ego feature scores, as the \gls{sae} might be able to extract the most important features, but considerably deteriorates in the feature disentanglement.
For our selected \gls{sae}, we aim to choose the latent space dimensionality high enough that individually important features can develop into independent components, but low enough, so that memorization effects and too strong feature dispersion are prevented.

% \subsection{Interpreting Neurons / Identifying Components of Driving Decisions}
\subsection{Understanding Concepts of Driving Decisions}
\label{sec:interpneuron}

\begin{figure*}
    \centering
    \includegraphics[width=\linewidth]{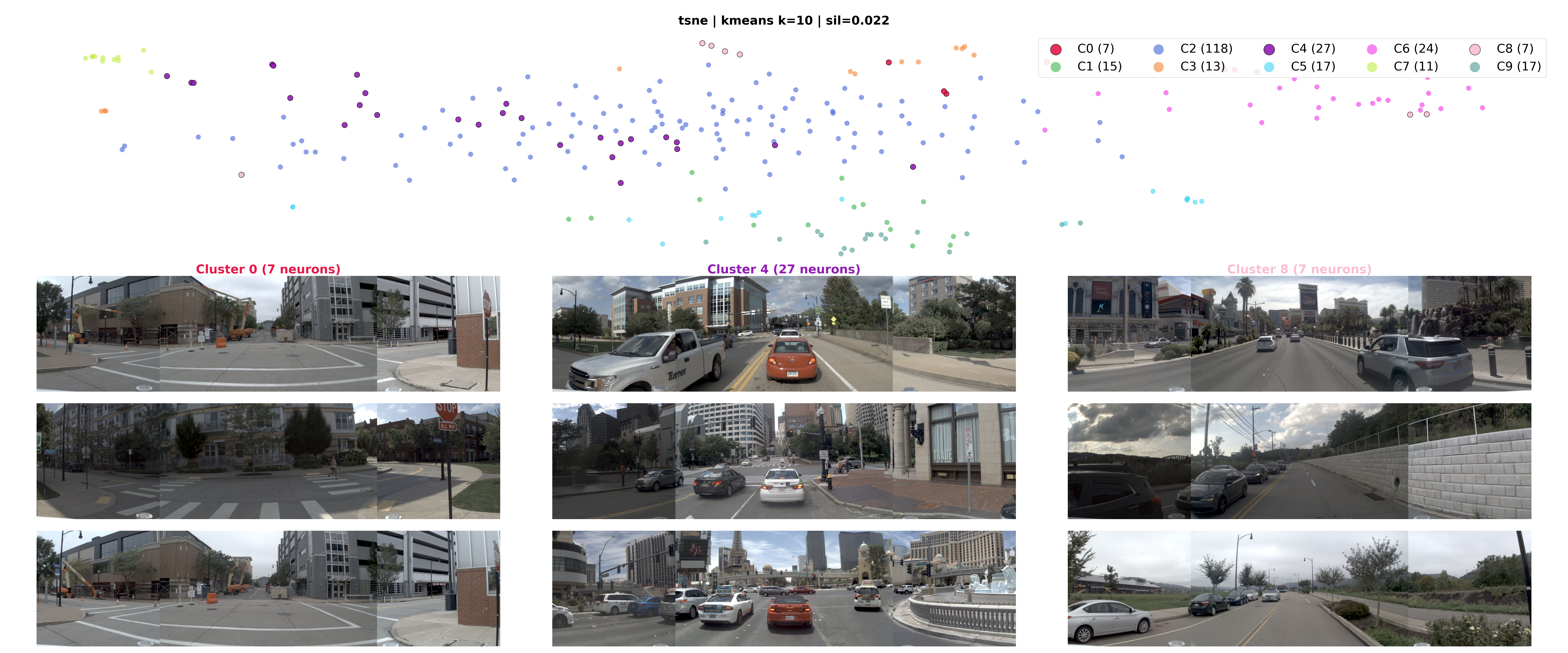}
    \caption{Overview of the neuron clustering: The \gls{sae} neurons form a distribution of individual concept directions with semantically similar neurons assembling together (top). By visualizing the top-activating samples per cluster (bottom), semantic meaning can be assigned as described in Section \ref{sec:interpneuron}.}
    \label{fig:cluster_overview}
\end{figure*}

A crucial part in using \glspl{sae} is the interpretation of neurons and feature directions, which is the third step of our framework from Section~\ref{sec:semantics}. 
While \glspl{sae} promise an increase of interpretability in the latent space, they do not guarantee that all neurons correspond to well-defined concepts.
Consequently, constructing an exhaustive dictionary that assigns clear semantics to every neuron in the context of the application is generally not feasible.
However, there are various strategies for extracting valuable insights into the latent space.
We propose the following methodologies for assigning semantic meaning to neurons, that can be used interchangeably or in combination.

By \emph{Neuron Maximization}~\cite{olah_feature_2017,achtibat_attribution_2023} we are able to extract the samples in the reference data, which maximally activate a chosen SAE neuron in the prediction of a trajectory. 
Through this, we reveal patterns in the input data causing these extreme activations (Fig.~\ref{fig:cluster_overview} bottom) and, thus, we are able to relate these patterns to the concepts in the model's feature space that correspond to the chosen SAE neuron. 
We directly combine the maximization with clustering neurons to group similar patterns and label them more efficiently.

\emph{Neuron Clustering} uses \gls{sae} activations on a reference dataset to compute similarities between the \gls{sae} neurons. 
Instead of clustering for the model understanding of the traffic scenes, the latent dimension of the \gls{sae} is clustered, thereby grouping neurons with similar activation behavior.
This reveals shared characteristics between neurons and allows for assigning semantic meaning to neuron groups.
We observe that multiple neurons activate for cars located in front of the ego vehicle, where the neurons differ in their sensitivity based on the distance to the vehicle, the type, and the color.

Fig.~\ref{fig:cluster_overview} visualizes the clustering of \gls{sae} neurons together with the samples maximally activating the neurons assigned to the respective cluster.
We observe distinct clusters corresponding to, for example, intersections and lane markings (cluster 0), following other traffic participants (cluster 4), and driving on straight roads (cluster 8).
These clusters, together with their associated neuron encodings, provide a basis for further analysis of the model’s scene understanding—for instance, by monitoring neuron activations under controlled input variations.
Moreover, they enable investigating how groups of neurons influence the prediction heads of the end-to-end model, as will be explored in Section~\ref{sec:resultcircuit}.

In the \emph{Neuron Attribution} (Section~\ref{sec:saeattr}), we compute the relevance for a selected trajectory---typically the predicted one by the model---towards the input. 
Hereby, we condition the relevance propagation on one or multiple \gls{sae} neurons, as defined by CRP~\cite{achtibat_attribution_2023}.
The resulting relevance in the camera input highlights which parts of the input were utilized by the selected neurons when scoring the trajectory, directly showing us the information exciting the respective neuron. 
We use the relevance maps, which are exemplary shown in Fig.~\ref{fig:sample_composition}, as additional evidence for assigning semantic concepts to neurons.
However, the relevance maps may be deceptive due to multiple reasons.
First, individual neurons may encode distributed features corresponding to scene composition rather than localized concepts.
They may also encode concepts like hue, lighting or weather, which is difficult to display with heatmaps; or relate to "elsewhere", highlighting everything but a specific concept, given that the concept is present~\cite{fel_rabbit_2025}.
Finally, the neuron may rather relate to other inputs like the ego status, or the semantic signal from the neuron may be superimposed in the heatmap by network effects.
Especially for the GTRS, the construction of the image backbone with the wide-range image input and the usage of convolutional layers facilitates the propagation of spurious features by amplifying overlaying stride signals in the backward pass, causing high attribution in the image center possibly superimposing the semantic signal.

We use \emph{Input Perturbation} to analyze how neuron activations change when specific input signals are removed.
In the end-to-end setting, systematically removing individual camera inputs or ego-state variables allows us to identify which neurons are most affected by each input modality. 
For the iPAD model, we observe that removing the backward-facing camera does not alter neuron activations, calling its contribution to the overall prediction into question.
Generally, we find no measurable gradient from the model prediction to the backward camera, suggesting that the camera is not required for optimizing the model on the given training data.
Consequently, the backward-facing camera could either be omitted or requires explicit mechanisms during training to enforce its utilization.
This behavior may extend more broadly to models trained on the \gls{navsim} dataset.

\subsection{Understanding the Composition of a Scenario}

\begin{figure*}
    \centering
    \includegraphics[width=\linewidth]{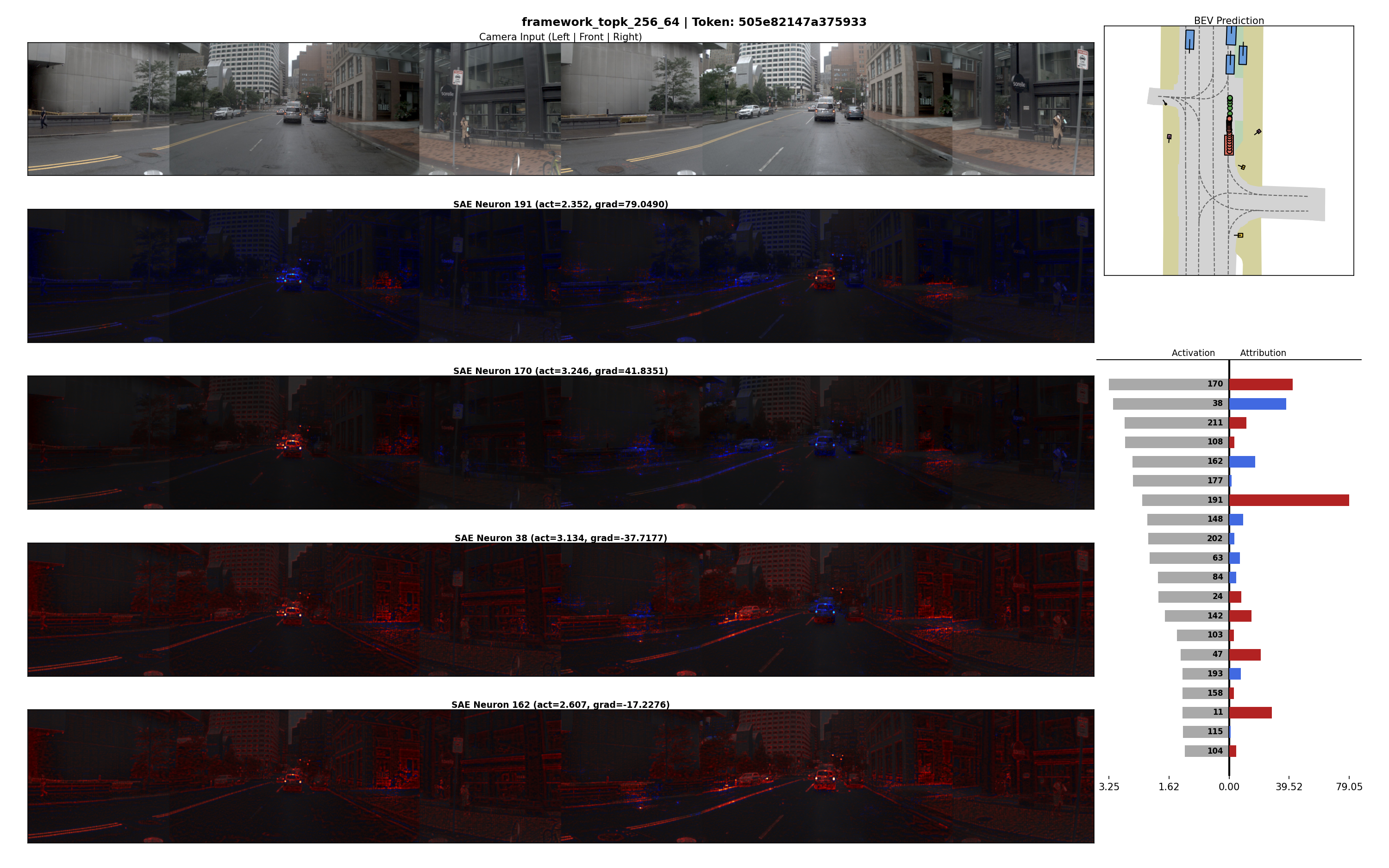}
    \caption{Composition of concepts in a single sample: The top row shows the camera input and, on the right side, the BEV visualization with ground truth (green) and predicted (red) trajectory. Below, the \gls{sae} neuron attribution towards the camera input is shown for the top-4 neurons. Red indicates positively contributing pixels and and blue indicates negatively contributing pixels, with black being neutral. An overview of the top-activating neurons with their attribution is depicted in the bar-chart on the right.}
    \label{fig:sample_composition}
\end{figure*}

With our proposed integration of an \gls{sae} into the model structure, single scenes and model decisions can be further analyzed based on \emph{what is present} in a scene (feature activation) and \emph{why it matters} for the driving decision (feature attribution).
The framework thus provides a stronger level of interpretability not only for scenes in the dataset, but also for different trajectory proposals within a scene, as the \gls{sae}-based concept decomposition runs on each trajectory proposal individually.

Fig.~\ref{fig:sample_composition} shows the camera input of a single sample (top), the referring BEV visualization with prediction and ground truth (top right), the relevant \gls{sae} neuron activations and attributions (right), and the camera attribution for the four most important neurons according to a mixture of activation and attribution.

In the shown example, the rankings of top-activating and top-attributed neurons differ significantly highlighting their differing purpose.
A highly activated neuron may not necessarily be used for the prediction of the final trajectory.
While the neuron activation refer to the presence or recognition of features in the scene, the attribution rather highlights positive and negative influence on the selected trajectory representing how the information has been used specifically.
Other trajectories may induce considerably other attribution scores matching the relation between \gls{sae} features and the driving maneuver.

For an even finer distinction, the prediction heads of the GTRS model can be individually addressed to compute the attribution of neurons for single \gls{pdm} components.
For the selected scene, we select the most important neurons by computing the product of activation and attribution and observe the following:
Neurons 191 and 170 encode for traffic participants infront of the ego vehicle.
Interestingly, they highlight the ahead-driving car at the two different time steps, which are joined in the camera input as the left and right side of the image.
The difference in the neurons and their attribution hints at a temporal distinction in the recognition of traffic participants.
Neuron 38 generally describes the in-front-car on road layouts similar to the one in this example, while neuron 162 refers to yellow road markings at left turns. 
The high activation of this neuron suggests the general possibility of turning left as illustrated by the BEV visualization. However, the attribution for this neuron is logically negative for the predicted trajectory since  going straight scores higher.

\subsection{Connecting Neurons with the Trajectory Scoring}
\label{sec:resultcircuit}
% \subsection{Verifying Circuits for single Trajectory Heads}

Having investigated the options for extracting and interpreting meaningful features from the latent space and decomposing a single sample into its concepts, we take a step further to gain an even deeper understanding of the model.
In the fourth step (Section~\ref{sec:circuits}), we determine how the \gls{sae} neurons influence the individual prediction subscores of the \gls{pdm} score.
The \gls{pdm} score is constructed from multiple attributes that are often predicted independently by the models. 
These attributes include progress along the ego vehicle's future trajectory, lane keeping, and traffic light compliance, which improve the score, while collisions caused by the ego vehicle decrease it.
Specifically, the GTRS model contains a dedicated head for each attribute, enabling a head-specific evaluation of the model's functionality.

\begin{figure}
    \centering
    \includegraphics[width=\linewidth]{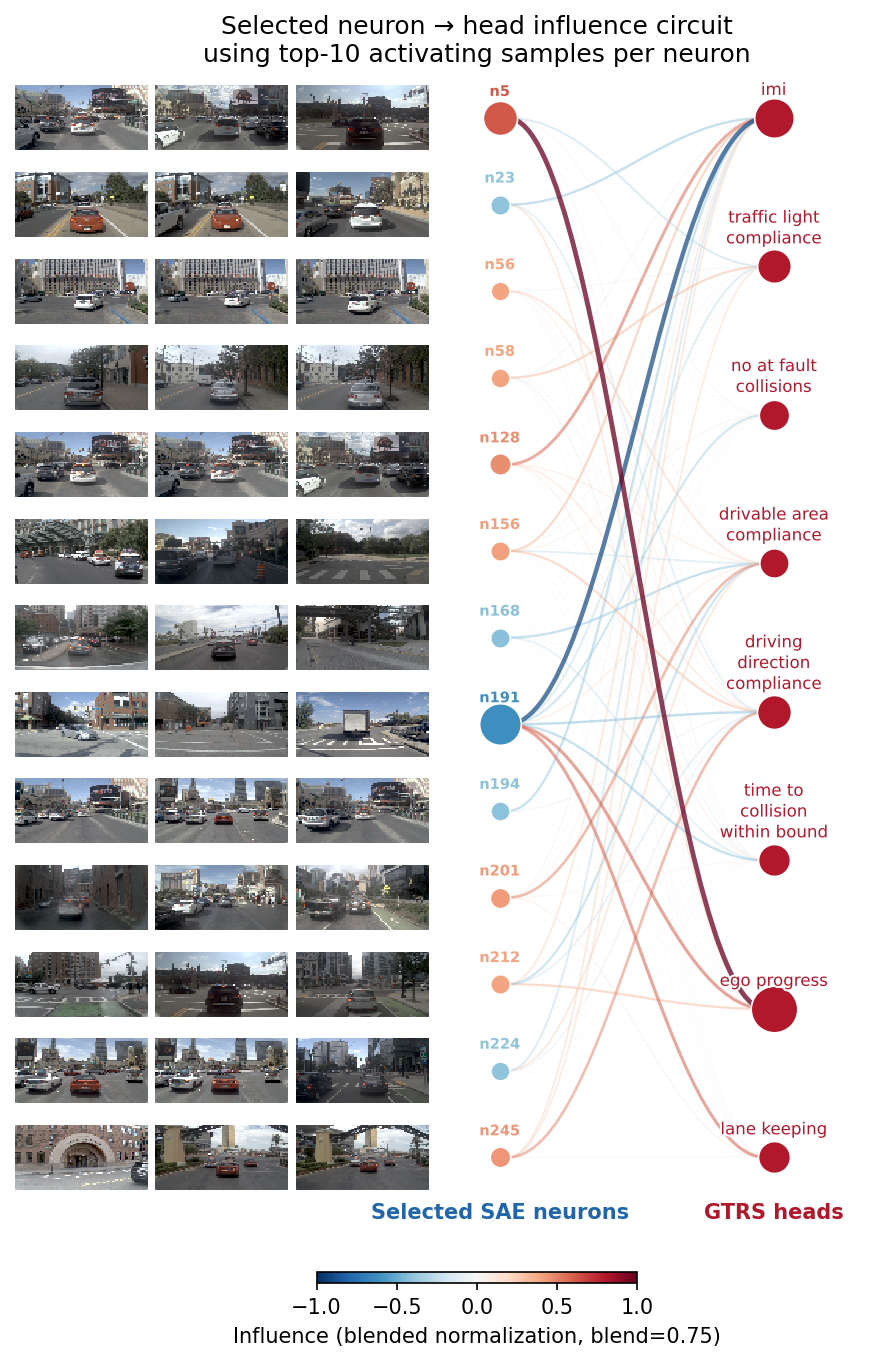}
    \caption{Circuit Visualization from \gls{sae} Neurons associated with "Following a Car Driving Ahead" to the Prediction Heads of the GTRS Model.}
    \label{fig:circuit_car_ahead}
\end{figure}

Circuit analysis enables the extraction of subgraphs that describe specific functionalities of the model.
We combine circuit analysis with \glspl{sae}~\cite{marks_sparse_2025} in order to connect the induced \gls{sae} directly to the predicted scoring heads, allowing us to explicitly evaluate the influence of \gls{sae} neurons on the prediction heads.

From our neuron clustering in Section~\ref{sec:interpneuron}, we select the cluster with neurons that broadly activate for scenarios in which a vehicle is positioned ahead of the ego vehicle.
Fig.~\ref{fig:circuit_car_ahead} shows the circuit visualization connecting these neurons to the prediction heads of the GTRS model.
The 10 top-activating samples per neuron have been used for the circuit computation.
While a car in front may naively just be associated with collision-based scores, the circuit clearly displays that all scoring heads are significantly triggered by the selected neurons.
Naturally, the ego progress is affected by the lead vehicle being the biggest constraint on how far the ego vehicle can advance.
The drivable area and lane keeping scores are accountable, as the front vehicle naturally shows a safe trajectory, or when lane changes for overtaking are considered.
For the traffic light compliance, a learned correlation may associate a stopping lead vehicle with the appearance of red lights or intersections requiring the car to slow down.

Interestingly, the \gls{sae} neurons related to "car ahead" scenarios seem to differ in how they affect the scoring heads.
We assume that there are differing conceptions of the situation following a leading vehicle in the model that encode partially redundant information, but are distinguishable in how they are used by the scoring heads.
Each scoring head learns a partially distinct subgraph, with some \gls{sae} neurons shared across heads while others contribute to head-specific structures.

As GTRS evaluates all trajectories in its predefined vocabulary, both prediction scores and corresponding \gls{sae} activations can be computed for any candidate trajectory.
In the previous analysis, we focused exclusively on the top-scoring trajectories, which naturally avoid collisions with the leading vehicle.
However, recomputing the circuit for forward trajectories that are likely to result in a collision reveals substantially different influencing signals.
In particular, potentially colliding trajectories induce stronger activations in the collision-related heads, reflecting a substantially increased influence of the "car ahead" neurons.
Specifically, when more risky trajectories that provoke collisions are selected, the "car ahead" neurons exert a substantially stronger influence on the activation of the time-to-collision-within-bound and no-at-fault-collisions heads, validating the model's intended behavior.

The circuit analysis further reveals a limitation of the overall learning scheme, as influencing factors in the decision-making process are significantly shaped by correlations that, while intuitively plausible, are not causally grounded. 
For instance, the observed influence of the "car ahead" neuron 58 from Fig.~\ref{fig:circuit_car_ahead} on the traffic light compliance score reflects co-occurring patterns in the data.
A vehicle closely located in front of the ego vehicle often indicates a traffic situation involving an intersection or a traffic light where the ego vehicle closely approaches the preceding vehicle.
Consequently, the model associates the presence of a close leading vehicle with traffic light compliance, exploiting a statistical correlation rather than only relying on the causal factor---the actual presence of a traffic light.

% \subsection{A remark on activations and attributing inputs via the GTRS scoring functions}
\subsection{A Remark on Model Construction and Interpretability}

The investigated models, iPAD and GTRS, are both predicting the \gls{pdm} score on a set trajectories to select the best fitting one for the future.
Interestingly, the \gls{pdm} score covers multiple different characteristics of a valid trajectory incorporating the driving direction, traffic laws and collision avoidance.
While the scoring function of the iPAD focuses on the overall \gls{pdm} prediction, the GTRS model predicts the single subscores with individual heads.
By design, splitting the subscores into separate parts allows for an individual inspection regarding the single characteristic, enabling a better interpretability.
Especially in combination with our approach, the scoring of single heads can be attributed more easily to the \gls{sae} neurons, allowing for a more thorough inspection.

When handling attributions from the scoring heads to the \gls{sae} neurons, an interesting contradiction arises.
The GTRS heads directly reproduce the construction of the \gls{pdm} score, providing one prediction head per subscore.
While some heads like the ego progress are intuitively interpretable and contribute positively towards the score, other heads capture compliance with driving rules as traffic light compliance, drivable area compliance and time to collision within bound.
These heads effectively predict the absence of a violation rather than the presence of clues.
Interesting samples for inspection are in these cases not the top-activating samples, but the ones with the lowest scores.
While vision-based AI models like object detection models are activation-based with e.g. convolutional filters activating on detected patterns, the scoring functionality of the penalty heads is rather unintuitive.
A high score rather corresponds to the absence of evidence for a violation than the presence of evidence for compliance.
For instance the traffic light compliance head currently yields a maximum score, if there exists no traffic light in the scene.

This construction of the scoring heads collides with the principles of activation-based AI models impeding especially the use of attribution-based methods for interpretability, as they typically extract high-activating signals best.
In the construction of future models, we suggest to incorporate our thoughts on interpretability and explore the activation-based approach of scoring violations and compensating for the inversion in the post-processing.

% \subsection{Manipulation}
\subsection{Manipulating Driving Decisions}

Beyond a lense into the model's functioning and single decision processes, integrating an \gls{sae} provides us the opportunity to modify the learned feature influence and thereby achieve an improved closed-loop behavior.
As the last step of our framework (Section~\ref{sec:manipulation}), we perform a targeted manipulation by masking the activation of selected \gls{sae} neurons and re-evaluate the predicted trajectory on the manipulated reconstructions.
Note that the evaluation is based on the objective scoring instance, rather than the model’s predicted scores.

Table~\ref{tab:manipulation} summarizes the aggregated scores on the \gls{navsim} evaluation dataset, when our manipulation is applied.
We select three neurons based on our circuit-level analysis.
In particular, neuron $177$ is identified having a negative influence across multiple scoring heads, suggesting potentially malicious behavior.
Both neurons $59$ and $71$ show a close correlation to neuron $177$, while also having remarkable influence on multiple heads.
By removing only the three selected neurons $59, 71$ and $177$, the overall \gls{epdms} score of the model improves significantly by $0.096$ points compared to the GTRS version with the unchanged \gls{sae} included.
Since the injection of the \gls{sae} slightly decreases the performance, we additionally compare to the original baseline model without the \gls{sae}.
Notably, the manipulated model surpasses the baseline, achieving an \gls{epdms} of $0.5926$ compared to $0.524$.
Investigating the individual components of the score, we measure improvements in nearly all driving metrics.
Especially the scores for drivable area compliance and driving direction improve substantially.
The collision-based scores improve slightly less, which may also relate to some extent to the manipulated model rather staying within the drivable area.
The only score that is negatively affected is the ego progress hinting at the general behavior of the car to be more conservative and trade rapid progress for increased safety.

\begin{table}[t]
\centering
\caption{Effect of combined zero-ablation of \gls{sae} neurons $\{177, 59, 71\}$ on
the \gls{epdms} score and its sub-metrics. ``Baseline'' denotes the model with unmodified \gls{sae},
``Ablated'' the manipulated model, and $\Delta$ their difference.
Positive values indicate improvement.
Original \gls{epdms} without \gls{sae}: 0.524}
\label{tab:manipulation}
\renewcommand{\arraystretch}{1.15}
\setlength{\tabcolsep}{6pt}
\begin{tabular}{lccc}
\toprule
\textbf{Metric} & \textbf{Baseline} & \textbf{Ablated} & \boldmath$\Delta$ \\
\midrule
\gls{epdms} score            & 0.4963 & 0.5926 & $+0.0963$ \\
\gls{pdm} score              & 0.6047 & 0.6897 & $+0.0850$ \\
Ego progress                 & 0.6378 & 0.5378 & $-0.1000$ \\
Time-to-collision w/i bound  & 0.9004 & 0.9667 & $+0.0663$ \\
No at-fault collisions       & 0.9179 & 0.9722 & $+0.0543$ \\
Drivable area compliance     & 0.7457 & 0.8754 & $+0.1298$ \\
Driving direction compliance & 0.8943 & 0.9645 & $+0.0702$ \\
Traffic light compliance     & 0.9987 & 0.9989 & $+0.0002$ \\
\bottomrule
\end{tabular}
\end{table}

\begin{figure}[ht]
    \centering
    \includegraphics[width=0.9\linewidth]{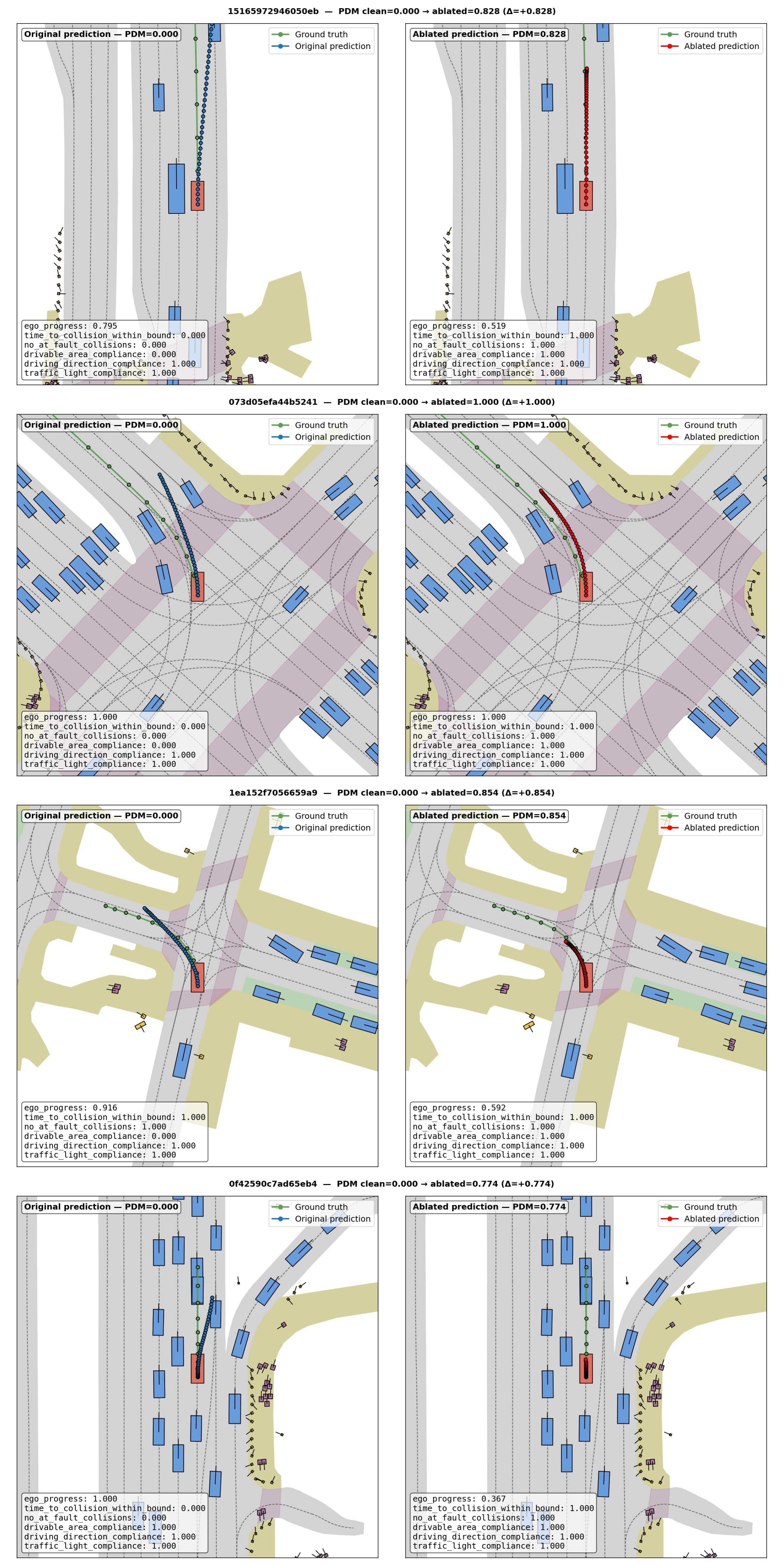}
    \caption{Comparison of predictions from the original and the manipulated model in selected examples.}
    \label{fig:intervention_examples}
\end{figure}

Inspecting the comparative examples in Fig.~\ref{fig:intervention_examples}, the assumed change of behavior in the model is affirmed, as risky trajectories with high progress are replaced by more moderate trajectories that favor the current driving lane instead of maximal progress.

\section{CONCLUSION}
End-to-end autonomous driving models are increasingly approaching real-world deployment; however, their lack of interpretability remains a critical challenge for model validation, profound error analysis, and targeted improvement.
In this work, we introduce a thorough, comprehensive framework for leveraging interpretability in such models by enabling access to their latent representations and explicitly linking learned concepts to prediction scores.
We show how human-understandable concepts can be extracted, analyzed and how they interoperate with the different prediction heads for selecting future trajectories.
Through this analysis, we provide a means to validate whether the model bases its decisions on meaningful semantic concepts and reasoning processes.
We reveal how the predictions of end-to-end autonomous driving models are frequently affected by learned correlations that do not necessarily reflect causal coherence, thereby facilitating a deeper understanding or model failures and undesired behaviors.
Finally, we show how to translate insights from the previous analyses into targeted interventions by providing an approach to easily adapt model inference at test time without retraining.
The modified model exhibits a clear performance gain in the \gls{epdms}, replacing risky progress-oriented behavior with safer and more compliant trajectories.
Illustrating how interpretability can support in model validation, debugging, and refinement, we believe this work represents a step toward more transparent, trustworthy, and controllable autonomous driving systems.
Concurrently, our results highlight the need for further research into the discovery, characterization, and causal understanding of learned concepts in autonomous driving models. 
While the proposed framework provides a foundation for concept-based analysis and intervention, fully understanding the semantic concepts and their interaction in the decision-making process remains a largely open research problem.

\section*{Acknowledgments}
The research leading to these results is funded by the German Federal Ministry for Economic Affairs and Energy within the project “Safe AI Engineering–Sicherheitsargumentation befähigendes AI Engineering über den gesamten Lebenszyklus einer KI-Funktion". The authors would like to thank the consortium for the successful cooperation.

\bibliography{ijcai25}
\bibliographystyle{IEEEtran}

\clearpage

\end{document}